%% file: main.tex
\documentclass{article}
\PassOptionsToPackage{numbers, compress}{natbib}

\usepackage[main,final]{neurips_2026}
\usepackage{amsmath}
\usepackage{amssymb}
\usepackage{mathtools}
\usepackage{amsthm}
\usepackage{multirow}
\usepackage{mathabx}
\usepackage{amsfonts}
\usepackage{wrapfig}
\usepackage{fontawesome5}
\usepackage{xcolor}

\definecolor{cbBlue}{HTML}{0072B2}
\definecolor{cbOrange}{HTML}{E69F00}
\definecolor{cbGreen}{HTML}{009E73}
\definecolor{cbPurple}{HTML}{CC79A7}
\definecolor{cbVermilion}{HTML}{D55E00}

\usepackage[utf8]{inputenc}
\usepackage[T1]{fontenc}
\usepackage{hyperref}
\usepackage{url}
\usepackage{booktabs}
\usepackage{capt-of}
\usepackage{amsfonts}
\usepackage{nicefrac}
\usepackage{microtype}
\usepackage{xcolor}
\usepackage{algorithm}
\usepackage{algpseudocode}
\usepackage{tcolorbox}
\newtcolorbox{promptbox}[1][]{
    colback=gray!5!white, colframe=gray!50!black,
    fontupper=\small\ttfamily, 
    arc=1mm, boxrule=0.5pt,
    title=#1,
    left=6pt, right=6pt, top=6pt, bottom=6pt
}
\usepackage{listings}
\definecolor{lstbg}{RGB}{248,248,248}
\definecolor{lstcomment}{RGB}{117,113,94}
\definecolor{lstkeyword}{RGB}{0,0,180}
\definecolor{lststring}{RGB}{163,21,21}

\lstdefinestyle{mintedlike}{
  backgroundcolor=\color{lstbg},
  basicstyle=\ttfamily\small,
  keywordstyle=\color{lstkeyword}\bfseries,
  commentstyle=\color{lstcomment}\itshape,
  stringstyle=\color{lststring},
  numberstyle=\tiny\color{gray},
  stepnumber=1,
  numbersep=6pt,
  framerule=0.2pt,
  breaklines=true,
  breakatwhitespace=true,
  showstringspaces=false,
  tabsize=1,
}

\title{They Can See It but Not Say It:\\Iterative Self Knowledge Re-expression \\ in Visual Reasoning Relative Pose Identification}

\author{%
  David S.~Hippocampus\thanks{Use footnote for providing further information
    about author (webpage, alternative address)---\emph{not} for acknowledging
    funding agencies.} \\
  Department of Computer Science\\
  Cranberry-Lemon University\\
  Pittsburgh, PA 15213 \\
  \texttt{hippo@cs.cranberry-lemon.edu} \\
}

\author{%
  Mengyu Wang$^{\text{\textcolor{cbBlue}{\faIcon[regular]{eye}}}\,\text{\textcolor{cbVermilion}{\faBinoculars}}}$ \And
  Ken Deng$^{\text{\textcolor{cbOrange}{\faIcon[regular]{eye}}}}$ \And
  Yftah Ziser$^{\text{\textcolor{cbGreen}{\faIcon[regular]{eye}}}}$ \AND
  Shay B. Cohen$^{\text{\textcolor{cbBlue}{\faIcon[regular]{eye}}}\,\text{\textcolor{cbPurple}{\faGlasses}}}$ \quad \quad
  Tiejun Ma$^{\text{\textcolor{cbBlue}{\faIcon[regular]{eye}}}\,\text{\textcolor{cbPurple}{\faGlasses}}}$ \\[1.5em]
  $^{\text{\textcolor{cbBlue}{\faIcon[regular]{eye}}}}$University of Edinburgh \quad
  $^{\text{\textcolor{cbOrange}{\faIcon[regular]{eye}}}}$University of Oxford \quad
  $^{\text{\textcolor{cbGreen}{\faIcon[regular]{eye}}}}$Nvidia Research
}

\begin{document}

\maketitle

\begingroup
\renewcommand{\thefootnote}{\textcolor{cbVermilion}{\faBinoculars} }
\footnotetext{Correspondence: \texttt{1306206684wmy@gmail.com}}
\endgroup

\begingroup
\renewcommand{\thefootnote}{\textcolor{cbPurple}{\faGlasses} }
\footnotetext{These authors jointly supervised this work}
\endgroup

\begin{abstract}
  The task of visual reasoning relative pose identification (VRRPI) has exposed critical weaknesses in the multi-view 3D spatial reasoning capabilities of modern vision-language models (VLMs). We present a systematic analysis of VLMs' behavior in VRRPI, uncovering stable output biases where VLMs disproportionately predict specific motion patterns regardless of visual evidence. To mitigate these inherent biases, we propose self knowledge re-expression with iterative debiasing (SKR-ID), an annotation-free pipeline that adapts VLMs for VRRPI using only randomly extracted unannotated image pairs. 
  SKR-ID transitions VLMs' output mechanism from generic next-token prediction to task-optimized classification heads. The pipeline proceeds in two stages: (1) a cold-start initialization using debiasing object-centric prompts to anchor motion inference in 2D visual evidence; and (2) a recursive logit-ranking mechanism that enforces statistical parity by using the median logit value as a dynamic threshold. 
  Experimental evaluation across frontier VLMs on standard benchmarks demonstrates that SKR-ID significantly mitigates systematic biases. Our method improves open-source VLMs' Macro-F1 scores from 35.0–45.4\% to 53.7–58.0\%, without using any annotated data. These results show that SKR-ID can effectively improve 3D reasoning performance from a severely collapsed baseline. The code is publicly available at \href{https://github.com/astudentuser/Codes_of_SKR-ID}{this repository}.
\end{abstract}

\section{Introduction}
VLMs have demonstrated exceptional ability in 2D perception and high-level semantic reasoning, excelling in tasks ranging from object detection~\cite{kuoopen,zhu2024survey} to complex image understanding~\cite{zhang2024vision}. However, these models are primarily optimized for 2D information processing, often lacking a fundamental grasp of 3D spatial geometry. This deficiency has been demonstrated in the task of visual reasoning relative pose identification (VRRPI), which requires inferring the relative view change between two camera viewpoints~\cite{deng2026lost}.

Despite the geometric clarity of VRRPI, recent benchmarks reveal that even frontier VLMs struggle to generalize beyond shallow 2D reasoning~\cite{du2024embspatial, stogiannidis2025mind,yang2025thinking,daxberger2025mm}. Advanced models often perform even below random chance, significantly lagging behind human performance. To investigate the cause of this failure, we conducted a systematic analysis of VLM behavior across various architectures. We uncover a robust and stable output bias: models exhibit a systematic preference for specific motion patterns or selections regardless of the actual visual parallax or prompt configuration. Our analysis reveals that these biases reflect frontier VLMs' inherent multi-view 3D spatial reasoning weaknesses, rather than being a surface-level artifact of instruction following. 

However, given the inherent correlation between 3D geometric transformations and the pixel-level shifts involved in 2D object detection and tracking, tasks where modern VLMs already demonstrate good performance~\cite{bao2025exploiting,sun2024chattracker,raja2025object}, we argue that these models are not inherently ignorant of 3D geometry. Instead, we believe some 3D spatial reasoning potential is latent within their parameters but remains underdeveloped because standard vision-encoder training often ignores 3D spatial information. This suggests that the performance gap is not only a lack of knowledge, but also a failure to correctly extract and express existing knowledge.

To use this latent potential, we build upon the framework of self-knowledge re-expression (SKR)~\cite{wang2026skr}. SKR posits that many deficiencies in LLMs are \emph{expression bottlenecks}, where the model possesses the requisite parametric knowledge but is constrained by the next-token prediction (NTP) output paradigm. By \emph{re-expressing} latent knowledge through task-optimized output heads, we can activate the model’s internal geometric reasoning without requiring human annotations or external solvers.

Motivated by this perspective, we propose the self knowledge re-expression with iterative debiasing (SKR-ID) method to adapt VLMs for VRRPI in an annotation-free setting. Our method relies solely on randomly extracted image pairs from unlabeled videos and proceeds in two stages. First, we use a debiasing prompt that guides the model to follow a reasoning chain, prioritizing 2D object-centric motion before inferring 3D camera-centric pose. Using this prompt for self-annotation, we implement the SKR process to transition the VLM from token generation to a task-specific classification head.

Second, we further address the model's inherent statistical bias through a logit-ranking mechanism. After the first round of SKR, the model provides task-specific prediction logits. We use these logits to re-calibrate the pseudo-labeling process. Instead of applying a fixed threshold, which would preserve the model's skewed distribution, we rank all self-annotations by their logits and set a dynamic threshold at the median. This ensures a balanced 50/50 label distribution for subsequent rounds of training, forcing the model to learn from balanced data and thereby mitigating bias in its spatial reasoning. This process of using SKR-adapted model to activate the next SKR iteration can be run recursively to further improve performance.

Our experimental results across three different frontier open-source VLMs, LLaVA-OneVision-7B, Qwen3-VL-8B, and VST-RL-7B, demonstrate that SKR-ID substantially reduces output collapse. The Macro-F1 scores are improved by 8.3--23.0 percentage points. Our detailed analysis also reveals that models adapted via SKR-ID successfully mitigate large-scale output biases.

In summary, our contributions are as follows:

\textbf{1. Analysis of Systematic Spatial Reasoning Biases in VLMs.} We provide a rigorous analysis of VLM behavior on VRRPI and identify inherent stable output biases that persist across model architectures and prompting strategies.
    
\textbf{2. The SKR-ID Method:} We propose an annotation-free method that enables VLMs to improve their 3D relative pose identification by re-expressing their own latent knowledge through task-specific heads and iterative logit-based debiasing.
    
\textbf{3. Performance Gains:} We demonstrate that SKR-ID consistently improves VRRPI performance across multiple frontier open-source VLMs,  substantially mitigating output imbalance by increasing Macro-F1 from 35.0–45.4\% to 53.7–58.0\%.

\section{Related Work}

\paragraph{Spatial grounding in VLMs}
Spatial grounding, binding language to locations, relations, and viewpoints, is a key stress-test for VLMs. Many benchmarks reveal systematic failures that persist across model scales and architectures~\cite{shiri2024empirical, huang2026spatialdiseunifiedbenchmarkevaluating, zhang2026revsi}.
Models that perform well on standard recognition tasks often drop to near-chance performance on cross-image correspondence and viewpoint-shifted localization~\cite{li2025viewspatial, yang2025mmsi, yin2025spatial, deng2026lost}.
Recent studies indicate that VLM spatial encodings lack invariance to viewpoint changes~\cite{kang2026linear}, leading to performance drops when reasoning moves beyond the egocentric frame~\cite{deng2026lost}. Existing solutions inject geometric priors or view-invariant objectives during pre-training~\cite{hong20233d, cheng2024spatialrgpt, zhang2026spatialstack, jiang2026spa3r}.
Our work identifies that VRRPI failures manifest as stable output biases.

\paragraph{3D spatial reasoning and relative camera pose}
Recovering relative camera pose from image pairs~\cite{hartley2003multiple} is a classical problem in computer vision. Traditional methods rely on establishing visual correspondences (e.g., keypoints or dense matches) and recover motion using epipolar geometry or known camera intrinsics~\cite{lowe2004distinctive, sun2021loftr, rao2026geometryawareepipolarcorrespondencegraphs}.
Recent work reformulates this problem in VLMs, where relative motion is expressed as discrete language descriptions~\cite{deng2026lost, yin2025spatial, yang2025mmsi, yeh2026seeing}. 
Unlike geometric methods, VLMs rely solely on image inputs and implicit reasoning, without explicit correspondence modeling or camera calibration. 
Existing approaches augment VLMs with depth estimation or 3D-aware representations~\cite{hong20233d, cheng2024spatialrgpt, qu2026loc3r, yang2026progressive}. Large-scale supervised datasets have also been used to inject 3D spatial knowledge~\cite{Chen_2024_CVPR, yang2025visual}.
However, a gap persists between semantic understanding and geometric reasoning. 
While VLMs can often identify shared objects across views, they still struggle to infer consistent view change~\cite{deng2026lost, zhang2026camcue}. 
We show that these limitations are driven by response biases that persist even after fine-tuning, motivating our annotation-free debiasing approach.

\paragraph{Self-training and knowledge re-expression}
Self-training uses model predictions as pseudo-labels for adaptation~\cite{asano2025selfiterativelabelrefinement, wang2026coft, wang2026skr}. Self-knowledge re-expression (SKR) further takes a complementary approach by replacing the autoregressive generation head with a task-specific expression head~\cite{wang2026skr}.
But they remain vulnerable to bias amplification when initial predictions are skewed.
Collaborative filtering and contrastive objectives have been proposed to suppress imbalanced labels in annotation-free settings~\cite{wang2022cmg, wang2026coft, asano2025selfiterativelabelrefinement}.
We extend SKR with an object-centric cold-start and a median-logit-ranking mechanism to enforce statistical parity, preventing distributional collapse during iterative refinement.

\section{Task Formulation}
\label{sec:task_forumulation}

\begin{wrapfigure}{r}{0.54\textwidth}
\centering
\vspace{-0.5cm}
\begin{center}
   \includegraphics[width=0.99\linewidth]{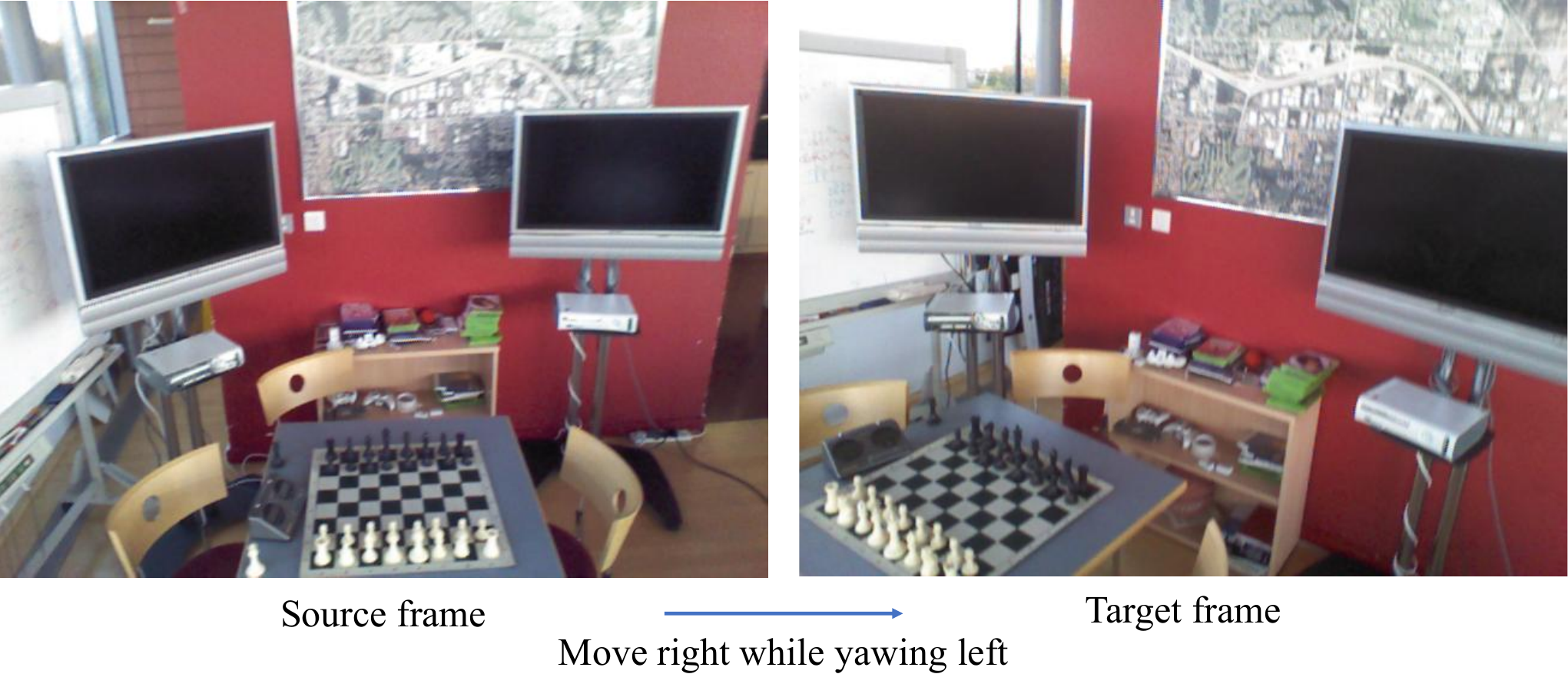}
    \vspace{-0.6cm}
   \caption{Example instance of the VRRPI task.}
\label{fig:knowledge_source_comparison}
\end{center}
\vspace{-0.3cm}
\end{wrapfigure}

We use the standard VRRPI-Bench dataset to develop experiments. It casts relative pose reasoning as a binary-choice visual question answering problem. Each example contains two views of the same scene, denoted by $(I_1, I_2)$, and asks the model to identify the camera motion that transforms the first view into the second. The benchmark discretizes this continuous transformation into two mutually exclusive natural-language options, ``\textit{Move left while yawing right}'' and ``\textit{Move right while yawing left}''. Although this task seems easy, studies have demonstrated that even frontier VLMs perform very poorly, with most VLMs performing near or even below random level~\cite{deng2026lost}.

Formally, using the annotations $\langle\textit{left}\rangle$ and $\langle\textit{right}\rangle$ to represent the two options ``\textit{Move left ...}'' and ``\textit{Move right ...}'' respectively, each instance is written as $x_i=(I^i_1, I^i_2, p)$, where $p = \text{prompt}(\langle\textit{left}\rangle, \langle\textit{right}\rangle)$ is the prompt template that includes the task instructions and the two options. Given a VLM $M$, the model is asked to output $\hat{y}_i \in \{\langle\textit{left}\rangle, \langle\textit{right}\rangle\}$ based on the input $x_i$ to identify the change in camera motion. 

\section{Bias Analysis of VLMs on VRRPI}
\label{sec:vlm_bias}

In this section, we conduct carefully designed evaluations on VRRPI-Bench to demonstrate that frontier VLMs do not fail this task randomly. Instead, they collapse onto a small set of motion descriptions, exhibiting a stable output bias.
Specific options are selected far more frequently than their symmetric counterparts, even when the input order is permuted. This phenomenon reveals that VLM failures in 3D spatial reasoning are driven by systematic response pathologies rather than stochastic noise, and these biases remain consistent across different models.

\subsection{Evaluation Design}

To characterize these biases, we evaluate each model under three controlled perturbations: (1) Base Inputs $(I_1, I_2, p)$, (2) Option-swap Inputs $(I_1, I_2, p_r)$, where $p_r=\text{prompt}(\langle\textit{right}\rangle, \langle\textit{left}\rangle)$ reverses the order of choices in $p$, and (3) Image-swap Inputs $(I_2, I_1, p)$, which exchanges the source and target frames. These settings isolate visual reasoning from the input configuration. In a well-calibrated model, swapping the option order should preserve the semantic answer, while swapping the image order should logically invert the answer because the underlying camera motion is reversed. Any persistent deviation from this behavior indicates that the model is relying on an indexing prior (a preference for a specific option position) or a motion prior (a preference for a specific movement type) rather than reasoning over the visual content.

Additionally, to verify whether the binary-choice constraint itself induces these failures, we introduce an extra option $\langle\textit{neither}\rangle :=$ ``\textit{Neither of them}'' and repeat the perturbation experiments. This serves as a control to ensure the model is not merely guessing due to forced choice.

\subsection{Problematic Outputs from VLMs}
We evaluate three frontier VLMs representing diverse architectures: LLaVA-OneVision, Qwen3-VL, and VST-RL (see Section~\ref{sec:model_baseline} for detailed descriptions).

\begin{table}[t]
    \centering
  \caption{Prediction ratios under controlled perturbations. Each cell reports the ratio of predicted labels $\langle\textit{left}\rangle$ vs. $\langle\textit{right}\rangle$. In all experimental settings, the ground-truth distribution is balanced (50.0\% / 50.0\%). Base, Swap Opt., and Swap Img. denote the base inputs, the option-swap inputs, and the image-swap inputs settings respectively.}
    \label{tab:bias-analysis}
        \input{tab/bias_analysis}
\end{table}

Table~\ref{tab:bias-analysis} reports the prediction ratios for $\hat{y}=\langle\textit{left}\rangle$ and $\hat{y}=\langle\textit{right}\rangle$ under the three settings. Because the benchmark is balanced by construction, a calibrated model should maintain an approximately 50.0\% / 50.0\% prediction distribution.

The results reveal a catastrophic breakdown in reasoning. LLaVA-OV exhibits a profound indexing bias, showing a tendency to always select the second option regardless of the option content or the input image order. This preference persists even after the introduction of a third option, confirming a complete collapse on the task.

Qwen3-VL and VST-RL exhibit answer biases, consistently preferring selection $\langle\textit{right}\rangle$ across all configurations. The addition of the $\langle\textit{neither}\rangle$ option does not solve this behavior, proving that these models are genuinely biased toward specific spatial outcomes rather than being constrained by the selection structure.

These problematic outputs may naturally lead to the assumption that frontier VLMs possess fundamental knowledge deficits in 3D spatial reasoning. However, given the inherent correlation between 3D geometric transformations and the 2D visual cues, which VLMs can understand correctly, it is unlikely that these models are entirely ignorant of 3D spatial principles.
Therefore, we posit that a primary cause of these failures is an expression bottleneck, where the model fails to correctly map its internal representations to the task's required outputs. This insight motivates our use of SKR-ID, an annotation-free method that relies on the models' own parametric knowledge to bridge this gap.

\section{Resolving the Bias: SKR with Iterative Debiasing}
\label{sec:skr_id}

We introduce SKR-ID, an annotation-free pipeline designed to transition VLMs from generic language generators into specialized view change reasoners. 
Our method extends SKR by formalizing the adaptation process as a recursive optimization loop that progressively decouples latent spatial knowledge from inherent model biases.

\subsection{Preliminaries}
The self knowledge re-expression (SKR) method is a fully local process that enables task adaptation using only unannotated data $\mathcal{D}_x$~\cite{wang2026skr}. The core premise of SKR is that LLMs/VLMs often possess much required parametric knowledge to solve a specific task $T$, but are constrained by the next-token prediction output paradigm. SKR decouples this latent knowledge from its token expression by transitioning the model from a generic mechanism $E_{\text{ntp}}$ to a task-specific expression $E_T$. The process executes in two steps: 

In the first step, \textbf{self-annotation}, the model uses its native generative mechanism, $E_{\text{ntp}}$, to perform ``knowledge extraction''. Given a task $T$ and unannotated inputs $\mathcal{D}_x$, the model generates textual outputs to complete the task and the outputs are processed into task-specific pseudo-labels $\hat{\mathcal{D}}_y$, such as numerical values or classification labels:
\begin{equation}
\label{eq:self_annotation} \mathcal{D}_x \underset{\text{Task Query}}{\xlongrightarrow{M_{\theta, E_{\textit{ntp}}}}} \mathcal{\hat{D}}_y,
\end{equation}
where $M_{\theta, E_{\text{ntp}}}$ represents the base model, with parameters $\theta$ and expression $E_{\text{ntp}}$, and $\mathcal{\hat{D}}_y$ represents the generated pseudo labels.

In the second step \textbf{knowledge re-expression}, the model is fine-tuned to map its internal hidden states directly to a task-adapted expression $E_T$, such as a classification head or regression layer. This bypasses the inefficiencies and errors of sequential token generation. This step is formalized as: 

\begin{equation}
M_{\theta, E_{\textit{ntp}}} \underset{(\mathcal{D}_x, \mathcal{\hat{D}}_y), \text{ Loss of }E_T}{\xlongrightarrow{\text{\quad\quad Fine-tune \quad\quad}}} M_{(\theta, \Delta \theta_T), E_T},
\end{equation}
where $\theta$ denotes the base parameters and $\Delta \theta_T$ represents the task-specific adapters and head.

\subsection{SKR-ID Stage 1: Cold-start via Debiasing Prompts}
Directly applying SKR to VRRPI is challenging because some base model's self-annotations $\hat{\mathcal{D}}_y$ are heavily biased, as discussed in Section~\ref{sec:vlm_bias}. However, we observe that while VLMs struggle with 3D pose reasoning, they excel at 2D object grounding and tracking.

We therefore design an object-centric reasoning prompt $p_{\textit{debias}}$ to anchor the self-annotation process in visual evidence. When the base model's outputs are highly biased, instead of directly querying the view change, the prompt guides the model through a causal chain: (i) identify a shared reference object in both frames, (ii) judge the object's movement, and (iii) infer the relative view change. This approach uses the model's 2D proficiency to generate more balanced labels.

It is important to note that this prompting strategy is not a direct solution to the task. Due to the models' inherent geometric reasoning weakness, performance under this setup is still not good enough, which will be discussed in the ablation study Section~\ref{sec:ablation}. 
Additionally, since SKR-ID uses only randomly sampled unannotated data for training, the object-centric prompt provides noisy, directionally informative pseudo-labels by grounding the initial prediction in the apparent motion of shared objects.
However, its value lies in providing a ``cold-start'' mechanism: it guides the VLM to generate a significantly more balanced distribution of initial pseudo-labels than direct querying would allow. This relative parity is essential to activate the SKR adaptation without immediately inheriting a collapsed output distribution.

Once these labels, represented by $\mathcal{\hat{D}}^0_y$, are generated, we implement SKR by replacing the model's final language modeling layer, which typically outputs a probability distribution over the entire vocabulary, with a task-specific binary classification head $E_{\textit{VRRPI}}$, whose outputs correspond directly to the two VRRPI motion classes. We train this head, together with the LoRA adapters, using the pseudo-labels and binary classification objective. Following the SKR framework, the whole process can be formalized as:

\begin{equation}
\label{eq:skr_process} 
\mathcal{D}_x \underset{p_{\textit{debias}}}{\xlongrightarrow{M_{\theta, E_{\textit{ntp}}}}} \mathcal{\hat{D}}^0_y,
\quad
M_{\theta, E_{\textit{ntp}}} \underset{(\mathcal{D}_x, \mathcal{\hat{D}}^0_y), \text{CE Loss}}{\xlongrightarrow{\text{\quad\quad Fine-tune \quad\quad}}} M^0_{(\theta, \Delta \theta^0_{\textit{VRRPI}}), E_{\textit{VRRPI}}}.
\end{equation}

The SKR model $M^0_{(\theta, \Delta \theta^0_{\textit{VRRPI}}), E_{\textit{VRRPI}}}$ now functions as a standard classifier, providing task-specific logits that represent the probability of the candidate view changes. In the following section, we represent the SKR-adapted models as $M^0_{(\theta, \Delta \theta^0_{\textit{VRRPI}})}$, omitting the explicit mechanism $E_{\textit{VRRPI}}$ as the architecture remains identical.

\subsection{SKR-ID Stage 2: Iterative SKR with Logit-based Debiasing}
The initial round of SKR yields the model $M^0_{(\theta, \Delta \theta^0_{\textit{VRRPI}})}$. It performs much better than the base model on VRRPI. Transitioning to a task-specific head enables $M^0$ to output continuous class logit distributions, which is far more informative than generated tokens. 
Both benefits suggest that $M^0$ has the potential to generate higher-quality self-annotations for the unannotated dataset $\mathcal{D}_x$, enabling a recursive process where each SKR model supports the next turn's SKR for better performance.

To implement this iteration, we define the turn index as $i$ ($i > 0$), where Stage 1 initializes the process with $M^0$. Using the class logits $L^{i}_x = \{\ell^i_1, \ell^i_2, \dots, \ell^i_n\}$ generated by the previous fine-tuned SKR model $M^{i-1}_{(\theta, \Delta \theta^{i-1}_{\textit{VRRPI}})}$, we implement a median logit-ranking strategy for self-annotation. We operate under a viewpoint symmetry prior: in a sufficiently large set of random changed viewpoints, the distribution of opposing directions (left vs. right) should be approximately uniform. Therefore, instead of using an absolute threshold 0.5 to decide labels, we can use the logits to force the self-annotations follow the balanced distribution. We rank the logits $L^i_x$ for the entire set $\mathcal{D}_x$ and define the threshold $T^i_{\text{med}}$ at the median value: $T^i_{\text{med}} = \text{median}(L^i_x)$. We then assign refined pseudo-labels $\hat{\mathcal{D}}^i_y$ by partitioning the data at this threshold. This ensures an exact 50/50 label distribution, effectively forcing the model to learn under a balanced supervision. Each iteration $i$ is formalized as:

\begin{equation}
\label{eq:iterative_process} 
\mathcal{D}_x {\xlongrightarrow{M^{i-1}_{(\theta, \Delta \theta^{i-1}_{\textit{VRRPI}})}}} L^i_x {\xlongrightarrow{T^i_{ \text{ med}}} \mathcal{\hat{D}}^i_y},
\quad
M_{\theta} \underset{(\mathcal{D}_x, \mathcal{\hat{D}}^i_y), \text{CE Loss}}{\xlongrightarrow{\text{\quad\quad Fine-tune \quad\quad}}} M^i_{(\theta, \Delta \theta^i_{\textit{VRRPI}})}.
\end{equation}

The iterated fine-tuned model $M^i_{(\theta, \Delta \theta^i_{\textit{VRRPI}})}$ demonstrates further improved performance and enhanced spatial reasoning.
The process terminates upon reaching a maximum iteration count or when the performance on the validation set (which is also split from the unannotated data) fails to improve for a set number of iterations. The complete pipeline is summarized in Algorithm~\ref{algorithm:SKR-ID}.

\begin{algorithm}
\caption{SKR-ID}
\begin{algorithmic}[1]

\State \textbf{Input:} Unlabeled image pairs $\mathcal{D}_x$, base VLM $M_\theta$ with parameters $\theta$, debiasing prompt $p_{\textit{debias}}$, max iterations $I_{\text{max}}$
\State \textbf{Output:} Optimized view change reasoner $M^i_{(\theta, \Delta \theta^i_{\textit{VRRPI}})}$
\State \textbf{Stage 1: Cold-start Initialization ($i=0$)}
\State $\hat{\mathcal{D}}^0_y \gets \text{Query } M_{\theta, E_{\text{ntp}}} \text{ on } \mathcal{D}_x \text{ using } p_{\textit{debias}}$
\State $M^0_{(\theta, \Delta \theta^0_{\textit{VRRPI}})} \gets \text{Fine-tune } M_\theta \text{ on } (\mathcal{D}_x, \hat{\mathcal{D}}^0_y) \text{ via cross-entropy Loss}$
\State \textbf{Stage 2: Iterative Logit-Ranking Refinement}
\For{$i = 1$ \textbf{to} $I_{\text{max}}$}
    \State Extract logits $L^i_x \gets M^{i-1}_{(\theta, \Delta \theta^{i-1}_{\textit{VRRPI}})}(\mathcal{D}_x)$
    \State Compute threshold $T^i_{\text{med}} \gets \text{median}(L^i_x)$
    \State Update pseudo-labels $\hat{\mathcal{D}}^i_y$: $\hat{y}_j^i = 1$ if $\ell_j^i \ge T^i_{\text{med}}$ else $0$
    \State $M^i_{(\theta, \Delta \theta^i_{\textit{VRRPI}})} \gets \text{Fine-tune } M_\theta \text{ on } (\mathcal{D}_x, \hat{\mathcal{D}}^i_y) \text{ via cross-entropy Loss}$
    \State \textbf{if} validation accuracy stagnates \textbf{then break}
\EndFor
\State \Return $M^i_{(\theta, \Delta \theta^i_{\textit{VRRPI}})}$

\end{algorithmic}
\label{algorithm:SKR-ID}
\end{algorithm}
\vspace{-0.2cm}

\section{Experimental Settings}
\label{sec:experiment_settings}

\subsection{Dataset and Evaluation}

\paragraph{Dataset}
We evaluate our methods on VRRPI-Bench~\cite{deng2026lost}, broken down by the 7 Scenes~\cite{1310glocker2013sevenscenes1,13shotton2013sevenscenes2} and ScanNet~\cite{dai2017scannet} subsets. Image pairs are grouped by approximate viewpoint separation into four difficulty levels: 15$^\circ$, 30$^\circ$, 45$^\circ$, and 60$^\circ$. The ground truth labels follow the binary classification objective detailed in Section~\ref{sec:task_forumulation}.
For SKR-ID, we use only random unlabeled image pairs sampled from ScanNet (no overlaps with VRRPI-Bench). The VRRPI-Bench dataset is used only for evaluation. This setup strictly follows an annotation-free protocol while evaluating on a standard labeled benchmark.

\paragraph{Metric}
We report Macro-F1 as the evaluation metric. Accuracy can be misleading when predictions are imbalanced. A model with heavily biased outputs can obtain seemingly high accuracy by outputting meaningless predictions. Macro-F1 treats classes equally by computing the F1-score for each class and averaging them: $\mathrm{Macro\text{-}F1} = \frac{1}{2}\left(\mathrm{F1}_{\langle\textit{left}\rangle} + \mathrm{F1}_{\langle\textit{right}\rangle}\right)$.  
For balanced binary labels, random guessing yields an expected Macro-F1 near 0.5. Heavily biased predictions result in scores below 0.5. Macro-F1 therefore serves as a better metric reflecting both accuracy and biases. All SKR-ID results are averaged over three runs, and the improvements are statistically significant under a t-test at the 0.5\% level.

\subsection{Models and Baselines}
\label{sec:model_baseline}
To evaluate the effectiveness of SKR-ID, we select three frontier VLMs as backbones:

\textbf{LLaVA-OneVision} (LLaVA-Onevision-Qwen2-7B-ov-hf)~\cite{lillava}: A unified framework designed for image and video understanding, recognized for its superior performance among open-source models.

\textbf{Qwen3-VL} (Qwen3-VL-8B-Instruct)~\cite{bai2025qwen3}: A recently released general-purpose VLM with enhanced perceptual capabilities, representing the state-of-the-art in open-source multimodal architectures.

\textbf{VST-RL} (VST-7B-RL)~\cite{yang2025visual}: A specialized model fine-tuned using the Visual Spatial Tuning framework, which specifically targets human-like spatial perception and reasoning.

These backbones represent a diverse spectrum of current multimodal research. By adapting these models via SKR-ID and comparing with them, we aim to demonstrate that even specialized spatial models struggle in the task of VRRPI and can benefit from our SKR-ID method.

\subsection{Implementation Details}
For model structure, we replace the vocabulary layer with a lightweight classification head attached to the last token’s final hidden state. This head consists of a LayerNorm and a two-layer MLP with a 1,024-dimension bottleneck and GELU activation. The final output is a scalar probability generated by a Sigmoid function, mapping internal representations directly to the target labels.

We apply parameter-efficient fine-tuning via LoRA on all LLMs' attention projection layers ($W_q, W_k, W_v, W_o$) with hyperparameters $r=4$, $\alpha=8$, and a dropout rate of $0.1$. Optimization is conducted using the AdamW optimizer (learning rate of $10^{-5}$) and a binary cross-entropy (BCE) loss objective.

We implement a recursive adaptation process with a maximum of 5 iterations. For the initialization stage ($i=0$), we use the object-centric debiasing prompt for LLaVA-OneVision and Qwen3-VL to address severe inherent biases, while VST-RL uses direct querying. Each iteration involves 10 epochs of training with a batch size of 8. Validation occurs every 200 steps to preserve optimal checkpoints.

All experiments are conducted using 4 NVIDIA A100 GPUs. The SKR training process uses 2 GPUs for LLaVA-OneVision and 4 GPUs for the other two models. Inference is performed on a single GPU for all models. Experimental prompts are detailed in Appendix~\ref{sec:prompt}.

\section{Results and Discussion}
\label{sec:results_discussions}
We introduce detailed experimental results and analysis in this section. Section~\ref{sec:main_results} presents the performance of the backbone models, representing state-of-the-art open-sourced VLMs, compared to our SKR-ID method. Section~\ref{sec:ablation} presents the performance of different stages of our SKR-ID method, analyzing the ablated contribution of our designs.
Section~\ref{sec:logit_debiasing} demonstrates the sensitivity of SKR-ID to different class-distribution priors, showing the effectiveness of median-thresholding.
Section~\ref{sec:confusion_matric} uses LLaVA-OneVision as an example to clearly demonstrate the change in confusion matrices across our pipeline, highlighting the systematic debiasing effect.

\subsection{VRRPI Results: Comparison with Backbones}
\label{sec:main_results}

\begin{table}
  \caption{Results (Macro-F1 in \%) on VRRPI-Bench.}
  \label{sample-table}
  \centering
  \begin{tabular}{llllll l@{\hspace{1pt}} lllll l@{\hspace{1pt}} l}
    \toprule
     \multirow{2}{*}{Model} & \multicolumn{5}{c}{7 Scenes} & & \multicolumn{5}{c}{ScanNet} & & All \\
     \cmidrule{2-6} \cmidrule{8-12} \cmidrule{14-14}
     & 15° & 30° & 45° & 60° & Avg. & & 15° & 30° & 45° & 60° & Avg. & & \\
     \cmidrule{2-14}
     LLaVA-OV & 37.6 & 33.9 & 44.7 & 35.8 & 37.6 &  & 38.0 & 36.6 & 31.0 & 33.9 & 34.8 &  & 35.0 \\
     \quad$\drsh$SKR-ID & 54.1 & 57.9 & 61.0 & 61.6 & 58.6 &  & 61.9 & 60.0 & 46.1 & 63.0 & 57.7 &  & \textbf{58.0}\\
     Qwen3-VL & 30.6 & 41.1 & 36.5 & 53.9 & 38.9 &  & 37.3 & 40.3 & 39.2 & 42.6 & 39.9 &  & 39.5 \\
     \quad$\drsh$SKR-ID & 42.7 & 52.1 & 58.0 & 64.0 & 54.2 &  & 36.0 & 47.3 & 65.8 & 67.9 & 54.2 &  & \textbf{53.8} \\
     VST-RL & 43.6 & 47.0 & 41.8 & 59.3 & 46.7 & & 37.2 & 35.5 & 46.6 & 57.6 & 44.2 & & 45.4 \\
     \quad$\drsh$SKR-ID & 47.7 & 52.4 & 53.2 & 61.5 & 53.7 &  & 43.5 & 52.7 & 52.6 & 69.4 & 54.5 &  & \textbf{53.7} \\
    \bottomrule
  \end{tabular}
  \label{tab:main_results}
  \vspace{-0.4cm}
\end{table}

Table~\ref{tab:main_results} summarizes the performance of frontier VLMs compared to those adapted via our SKR-ID method on the VRRPI-Bench. Appendix~\ref{app:complete_eval} reports the complementary evaluation metrics for these results, including accuracy, balanced accuracy, per-class F1, prediction ratios, and AUROC. The results demonstrate a consistent and significant performance improvement across all evaluated backbones. 

These base VLMs, due to their highly biased outputs on the VRRPI task, often fall below even the random-guess baseline (50.0\%). Following SKR-ID adaptation, all three models exhibit a Macro-F1 improvement of at least 8.3 percentage points. LLaVA-OV achieves the most dramatic recovery, with its average Macro-F1 score improving by 23.0 absolute points. Even for the specialized VST-RL model, which is pre-trained for spatial reasoning and initially shows a more balanced distribution, SKR-ID provides a clear boost from 45.4\% to 53.7\%. These substantial gains suggest that by transitioning from token prediction to a task-specific head and enforcing label balance, the model can successfully decouple its latent representations from its biased pre-training priors.

Overall, the consistent gains across different backbones validate SKR-ID as a robust, annotation-free method for adapting VLMs to the VRRPI tasks. The recovery of heavily biased models to much better performance confirms that current VLM spatial reasoning failures are related to the "expression bottlenecks" rather than a total absence of multi-view 3D spatial reasoning knowledge.

\subsection{Ablation Study: Contributions of Individual SKR-ID Components}
\label{sec:ablation}

\begin{figure}[tb]
\centering
   \includegraphics[width=0.98\linewidth]{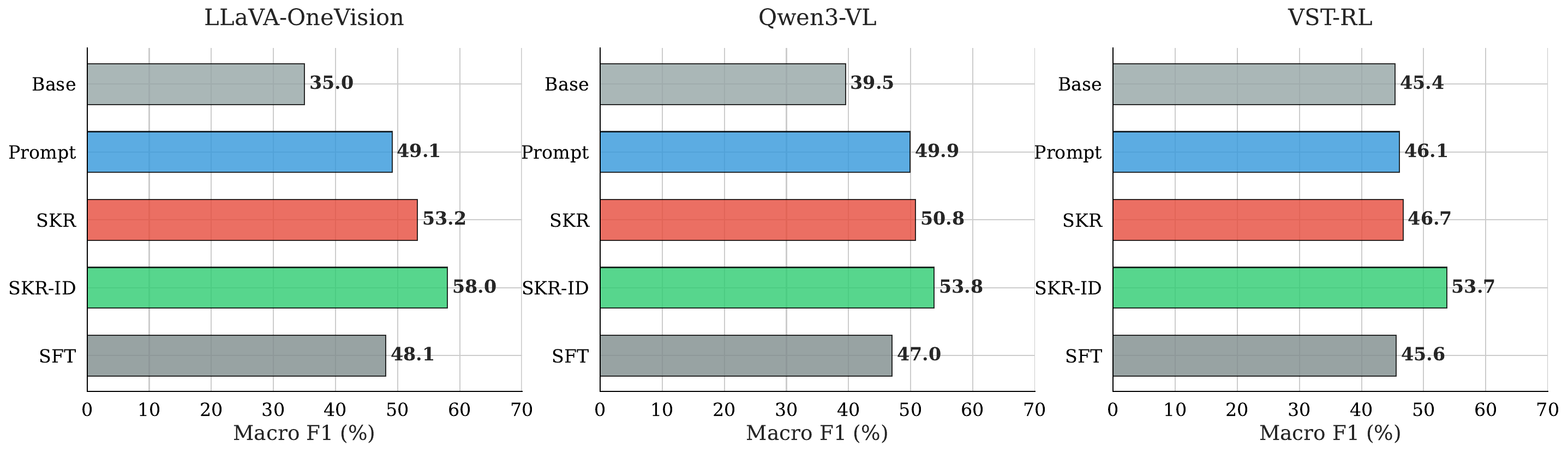}
   \vspace{-0.2cm}
   \caption{Comparison of VRRPI performance across adaptation phases, from base models to the final SKR-ID models. SFT denotes supervised fine-tuning using self-annotations from the base model.}
\label{fig:ablation}
\end{figure}

Figure~\ref{fig:ablation} illustrates the performance at each stage of the SKR-ID pipeline across the three backbone models. The ablation demonstrates that the transition from a base state to the final improved model is driven by the integration of object-centric prompting, knowledge re-expression, and recursive refinement. We also add a comparison with the standard supervised fine-tuning (SFT), which uses the self-annotations generated by the base model (the same as stage 1 initialization), but conducts SFT to fine-tune the base model rather than SKR. Appendix~\ref{app:detailed_ablation} provides the extended component ablation and additional baselines on LLaVA-OneVision.

The transition from base model to using special prompt (Stage 1 initialization) highlights the effectiveness of the object-centric prompt in debiasing VLMs' 3D spatial understanding. The performance of all three backbones improved after applying the debiasing reasoning. Models with the strongest initial bias benefited the most, LLaVA-OneVision jumped from 35.0\% to 49.1\%, demonstrating the prompt's critical role in anchoring 3D view change inference in 2D visual evidence.

The subsequent SKR method provides consistent improvements across all models. The first round of SKR, which re-expresses model outputs to classification information, further improves results. Importantly, the SKR models' inputs use the same direct query prompt as the base model, rather than the debiasing prompt. This confirms that the SKR mechanism can effectively unearth and formalize the model's latent spatial abilities that were originally only accessible through specialized prompting.

Finally, SKR-ID pushes all backbones to their peak performance (58.0\% for LLaVA-OneVision, 53.8\% for Qwen3-VL, and 53.7\% for VST-RL). By using predicted logits to enforce statistical parity through median thresholding, SKR-ID successfully unearths the models' potential for VRRPI.

Additionally, SFT results are clearly weaker than those of SKR-ID. While SFT outperforms the base model by partially mitigating biases via the debiasing prompt, it significantly underperforms the final SKR-ID. SFT remains constrained by the NTP paradigm and cannot provide task-specific information like class logits to support granular logit-ranking debiasing.

\subsection{Logit-Based Debiasing Analysis: Baselines and Prior Sensitivity}
\label{sec:logit_debiasing}
Stage 2 of SKR-ID uses median-based thresholding to mitigate output bias, based on the balanced class prior expected for randomly sampled pairs in our symmetric binary camera-motion task. To assess this design, we compare median thresholding across alternative base-model score constructions and examine its sensitivity to different class-distribution priors.

We first compare SKR-ID with median thresholding applied directly to scores from frozen VLMs, using three common score construction methods: (1) prompting the model to generate class probabilities, (2) aggregating token scores for class-relevant keywords, and (3) computing length-normalized sequence log-probabilities for the candidate answers. These controls use no task-specific classification head, fine-tuning, or iterative self-training. Table~\ref{tab:logit_debiasing} shows that median thresholding of these scores does not reproduce the performance of SKR-ID. Its effect varies by score construction and backbone, while SKR-ID performs best for all three models. This comparison indicates that changing the output balance alone is insufficient to obtain the gains of the full method.

We next examine the influence of the class-distribution prior. Because SKR-ID uses no annotated training data, we cannot construct a training set with a ground-truth class distribution. Therefore, we compare two approximate imbalance settings to assess the sensitivity of our prior: (1) constructing a 70/30 training subset using pseudo-labels from the object-centric prompt, and (2) replacing the median threshold with a 0.7-quantile threshold. The first setting does not guarantee a true 70/30 ground-truth distribution since the pseudo-labels may be noisy and some image pairs may be ambiguous. As shown in Table~\ref{tab:logit_debiasing}, the pseudo-label-based 70/30 setting moderately reduces Macro-F1 relative to standard SKR-ID, while the 0.7-quantile threshold causes a larger decrease. The median threshold performs best in these experiments, supporting the reasonableness of our balanced prior and the effectiveness of the method designed around this assumption.

\begin{table}[b]
\caption{Macro-F1 (\%) for median-threshold controls and class-distribution prior sensitivity on VRRPI. The 70/30 subset is constructed using noisy pseudo-labels from the object-centric prompt.}
\label{tab:logit_debiasing}
\centering
\small
\setlength{\tabcolsep}{4pt}
\begin{tabular}{lrrrrrrr}
\toprule
Model & Base & \shortstack{Prompt\\prob.} & \shortstack{Keyword\\token} & \shortstack{Sequence\\log-prob.} & \shortstack{70/30\\subset} & \shortstack{0.7-\\quantile} & Median \\
\midrule
LLaVA-OV & 35.0 & 32.3 & 34.3 & 43.4 & 56.7 & 47.2 & \textbf{58.0} \\
Qwen3-VL & 39.5 & 46.3 & 36.7 & 37.8 & 51.9 & 46.3 & \textbf{53.8} \\
VST-RL & 45.4 & 48.9 & 43.4 & 34.3 & 52.4 & 42.1 & \textbf{53.7} \\
\bottomrule
\end{tabular}
\end{table}

We also evaluate SKR-ID on a test subset with a 70/30 class distribution, with results reported in Appendix~\ref{app:imbalanced_eval}. These results further demonstrate SKR-ID’s effectiveness under the tested class-distribution shift, rather than only when the test and training data have similar class distributions.

\subsection{Bias Analysis: Evolution of the Confusion Matrices}
\label{sec:confusion_matric}

\begin{figure}[t]
\centering
   \includegraphics[width=0.96\linewidth]{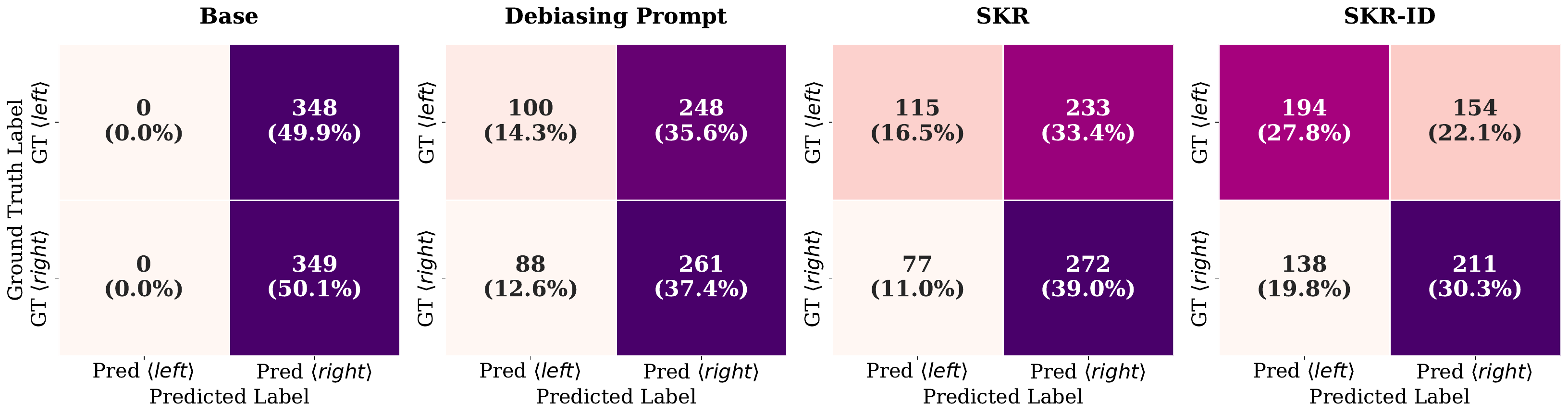}
   \vspace{-0.1cm}
   \caption{Confusion matrices change of LLaVA-OneVision model through the SKR-ID pipeline}
\label{fig:bias}
\vspace{-0.2cm}
\end{figure}

Figure~\ref{fig:bias} visualizes the evolution of prediction distributions for the LLaVA-OneVision model through each stage of the SKR-ID pipeline. This qualitative analysis reveals how the heavily biased outputs are progressively rectified.

As shown in the first matrix, the base model exhibits a total collapse of the output distribution. As analyzed in Section~\ref{sec:vlm_bias}, LLaVA-OneVision's direct outputs are heavily biased to label $\langle\textit{right}\rangle$. This confirms that the base model is not performing meaningful geometric reasoning but is instead trapped in a deep-seated language prior. The introduction of the object-centric debiasing prompt effectively activates the model's reasoning ability, anchored in 2D object tracking. While this breaks the initial collapse, expanding the prediction space to cover both labels, the results remain heavily skewed.

The transition to the recursive refinement loop marks a decisive change in performance. The first round of SKR (replacing the NTP head with a classification head) shows a clear trend of shifting the confusion matrix toward a diagonal structure. The following recursive SKR process, using logit-ranking debiasing, achieves a clear diagonal shape. This transition confirms that SKR-ID has successfully decoupled the model's visual representations from its pre-training textual biases, transforming the VLM into a specialized VRRPI identifier capable of reasoning 3D spatial change.

\section{Conclusions, Limitations, and Future Work}
\label{sec:conclusion}

We have demonstrated that VLMs possess latent spatial knowledge that is often obscured by systematic biases and the limitations of token generation. By using an iterative SKR method with median-logit ranking (SKR-ID), we successfully force the model to learn from a balanced representation of its own knowledge, effectively debiasing the spatial reasoning process for VRRPI. 

Despite the effectiveness of our method, our work has two primary limitations. First, we limited our evaluation to three specific backbones.  However, our selection includes the two best-performing models of differing architectures from the VRRPI benchmark and  the latest spatial-data fine-tuned VST-RL model. Therefore, they represent the competitive frontier of open-source VLM performance on this task. Second, the randomly extracted unannotated image pairs are heavily noised, including motion blur and geometric inconsistencies relative to the VRRPI-Bench standards. We intentionally retained these noises to prove that SKR-ID has minimal requirements for training data and that raw data can be extracted with marginal human effort. However, this also indicates significant potential for further performance improvement through including quality filtering.

This work opens a new perspective on VLM deficiencies by identifying stable output biases that likely affect a wide range of multimodal tasks beyond VRRPI. Our SKR-ID method shows strong potential for further improvement and for extension to a broader range of vision tasks.

\section*{Acknowledgments}

This work was supported by the UKRI Centre for Doctoral Training (CDT) in Natural Language Processing through UKRI grant EP/S022481/1.
We appreciate the comments and feedback by the reviewers and the area chair. We are grateful for funding provided by G-Research and St Catherine's College, Oxford to disseminate this study. We acknowledge the use of resources provided by the Isambard-AI National AI Research Resource (AIRR).
Isambard-AI is operated by the University of Bristol and is funded by the UK Government's Department for Science, Innovation and Technology (DSIT) via UK Research and Innovation; and the Science and Technology Facilities Council [ST/AIRR/I-A-I/1023]~\citep{mcintoshsmith2024isambardaileadershipclasssupercomputer}.
We also thank the Edinburgh Language Model team in their support.
The authors declare no competing interests.

\bibliographystyle{plainnat}
\bibliography{ref}

\appendix

\section{Prompts}
\label{sec:prompt}

We show the prompt variants mentioned in Section~\ref{sec:vlm_bias} (For each prompt template, we instruct VLMs in two ways: 1. to output the index of option; 2. to output the option itself. We report the best performance under these two output-formats.):

(1) Base inputs prompt (the same as image-swap inputs prompt):

\begin{promptbox}[Prompt for base inputs.]
\begin{lstlisting}
You are given two images from the same scene but different perspectives. The first image is the source image (src) and the second image is the target image (tgt). Based on these two images, determine if the camera movement from src to tgt is Move left while yawing right or Move right while yawing left. Please directly output your option. Do not output anything else.
\end{lstlisting}
\end{promptbox}

(2) Option-swap inputs prompt to test positional bias:

\begin{promptbox}[Prompt for option-swap inputs.]
\begin{lstlisting}
You are given two images from the same scene but different perspectives. The first image is the source image (src) and the second image is the target image (tgt). Based on these two images, determine if the camera movement from src to tgt is Move right while yawing left or Move left while yawing right. Please directly output your option. Do not output anything else.
\end{lstlisting}
\end{promptbox}

(3) Extended prompt with an additional $\langle neither \rangle$ option:

\begin{promptbox}[Prompt for base inputs (with extra option).]
\begin{lstlisting}
You are given two images from the same scene but different perspectives. The first image is the source image (src) and the second image is the target image (tgt). Based on these two images, determine if the camera movement from src to tgt is Move left while yawing right or Move right while yawing left or Neither of them. Please directly output your option. Do not output anything else.
\end{lstlisting}
\end{promptbox}

We also present the debiasing object-centric prompt here:
\begin{promptbox}[Debiasing object-centric prompt]
\begin{lstlisting}
You are given two images from the same scene but different perspectives. The first image is the source image (src) and the second image is the target image (tgt). First, identify a common object that appears in both images. Second, determine if this common object moved left or moved right from the src image to the tgt image.
\end{lstlisting}
\end{promptbox}

\section{Complete Evaluation Results on VRRPI-Bench}
\label{app:complete_eval}

Table~\ref{tab:complete_eval} supplements the Macro-F1 results in Table~\ref{tab:main_results} with accuracy (Acc), balanced accuracy (Bal Acc), per-class F1 (C1 F1, C2 F1), prediction ratios (C1 Ratio, C2 Ratio), and AUROC. The prediction ratios show the degree of output collapse in the base models. The balanced accuracy, per-class F1, and AUROC provide complementary measures of class discrimination. Across all three backbones, SKR-ID improves these measures and produces less collapsed predictions.

\begin{center}
\captionof{table}{Complete evaluation results on the balanced VRRPI-Bench test set. All metrics are reported as percentages.}
\label{tab:complete_eval}
\small
\begin{tabular}{llrrrrrrrr}
\toprule
Backbone & Method & F1 & Acc & \shortstack{Bal Acc} & \shortstack{C1 F1} & \shortstack{C2 F1} & \shortstack{C1 Ratio} & \shortstack{C2 Ratio} & AUROC \\
\midrule
\multirow{2}{*}{LLaVA-OV} & Base & 35.0 & 50.4 & 50.4 & 66.7 & 3.4 & 98.7 & 1.3 & 50.4 \\
& SKR-ID & \textbf{58.0} & \textbf{58.1} & \textbf{58.1} & 57.1 & 59.1 & 47.6 & 52.4 & \textbf{60.4} \\
\multirow{2}{*}{Qwen3-VL} & Base & 39.5 & 48.1 & 48.0 & 16.2 & 62.9 & 10.3 & 87.2 & 49.4 \\
& SKR-ID & \textbf{53.8} & \textbf{55.8} & \textbf{55.8} & 44.2 & 63.4 & 29.3 & 70.7 & \textbf{57.7} \\
\multirow{2}{*}{VST-RL} & Base & 45.4 & 46.6 & 46.6 & 37.4 & 53.5 & 35.3 & 64.7 & 46.6 \\
& SKR-ID & \textbf{53.7} & \textbf{54.1} & \textbf{54.1} & \textbf{49.2} & \textbf{58.1} & 40.5 & 59.5 & \textbf{57.3} \\
\bottomrule
\end{tabular}%
\end{center}

\section{Detailed Ablation Study of SKR-ID}
\label{app:detailed_ablation}

To study SKR-ID in greater detail, we conduct an extended ablation study on LLaVA-OneVision. We compare the following settings: \\
\noindent 1. \textbf{B}: Directly query the base model.\\
\noindent 2. \textbf{B + P}: Query the base model using the object-centric prompt.\\
\noindent 3. \textbf{B + P + SKR}: Use the object-centric prompt to generate pseudo-labels for one round of SKR.\\
\noindent 4. \textbf{B + P + SKR + I (without M)}: Iteratively apply SKR using the prompt-generated pseudo-labels, without median-based balancing.\\
\noindent 5. \textbf{B + SKR + M + I (without P)}: Initialize SKR using the base model's outputs, then iteratively refine the pseudo-labels using median-based balancing.\\
\noindent 6. \textbf{Full SKR-ID}: Use the object-centric prompt for initialization, followed by iterative SKR with median-based balancing.\\
\noindent 7. \textbf{Random guess}: Randomly predict one of the two selections.\\
\noindent 8. \textbf{Random guess SKR}: Initialize SKR with randomly generated balanced labels.\\
\noindent 9. \textbf{Random guess SKR-ID}: Initialize with randomly generated balanced labels, then apply the full iterative SKR-ID procedure.\\
\noindent 10. \textbf{Prompt SC3}: Apply self-consistency to three independently generated responses to the object-centric prompt.\\
\noindent 11. \textbf{Prompt SC5}: Apply self-consistency to five independently generated responses to the object-centric prompt.\\
\noindent 12. \textbf{Probing}: Train a probing layer on the model's final representations without adding adapters.\\
\noindent 13. \textbf{Probing with M + I}: Iteratively train the probing layer using median-based label balancing.

Results are reported in Table~\ref{tab:detailed_ablation}.
Iterative SKR without median thresholding gives only a limited improvement over one-round SKR, while omitting the object-centric prompt causes the initialization to collapse to one class. Random balanced labels can break this initial collapse, but random-label SKR and SKR-ID remain near the balanced-random result, indicating that class balance alone does not supply informative image-label correspondences. Prompt self-consistency provides only limited improvement over a single prompted response. The probing baselines also remain below the full SKR-ID result, including when combined with iterative median-based refinement.

\begin{minipage}{\textwidth}
\centering
\captionof{table}{Extended ablation study on LLaVA-OneVision. All metrics are reported as percentages. SC denotes self-consistency; probing trains a linear classification layer on the final multimodal representations without adapters.}
\label{tab:detailed_ablation}
\small
\setlength{\tabcolsep}{4pt}
\begin{tabular}{lrrrrrrrr}
\toprule
Setting & F1 & Acc & Bal Acc & C1 F1 & C2 F1 & C1 Ratio & C2 Ratio & AUROC \\
\midrule
\textbf{B} & 35.2 & 51.7 & 50.2 & 67.9 & 2.4 & 98.9 & 1.1 & 50.2 \\
\textbf{B + P} & 49.1 & 51.8 & 51.8 & 37.3 & 60.8 & 27.0 & 73.0 & 51.8 \\
\textbf{B + P + SKR} & 53.2 & 55.5 & 55.5 & 42.6 & 63.7 & 27.6 & 72.5 & 59.3 \\
\textbf{B + P + SKR + I (without M)} & 54.0 & 55.2 & 55.2 & 46.6 & 61.5 & 33.9 & 66.1 & 57.4 \\
\textbf{B + SKR + M + I (without P)} & 33.3 & 49.9 & 50.0 & 66.7 & 0.0 & 100.0 & 0.0 & 50.0 \\
\textbf{Full SKR-ID} & \textbf{58.0} & \textbf{58.1} & \textbf{58.1} & \textbf{57.1} & 59.1 & 47.6 & 52.4 & \textbf{60.4} \\
\textbf{Random guess} & 50.3 & 50.3 & 50.3 & 49.0 & 51.0 & 49.9 & 50.1 & 50.3 \\
\textbf{Random guess SKR} & 47.6 & 52.1 & 52.0 & 32.4 & 62.9 & 21.0 & 79.1 & 55.3 \\
\textbf{Random guess SKR-ID} & 50.8 & 51.7 & 51.6 & 44.5 & 57.2 & 37.2 & 62.8 & 51.6 \\
\textbf{Prompt SC3} & 49.6 & 50.5 & 50.5 & 56.4 & 42.8 & 64.0 & 36.0 & 50.5 \\
\textbf{Prompt SC5} & 49.8 & 50.8 & 50.8 & 56.7 & 42.9 & 63.8 & 36.2 & 50.8 \\
\textbf{Probing} & 34.2 & 50.2 & 50.2 & 1.7 & 66.7 & 0.7 & 99.3 & 60.5 \\
\textbf{Probing with M + I} & 50.2 & 52.1 & 52.1 & 40.6 & 59.9 & 30.7 & 69.3 & 54.6 \\
\bottomrule
\end{tabular}%
\end{minipage}

\section{Evaluation on Imbalanced Data}
\label{app:imbalanced_eval}

Table~\ref{tab:imbalanced_eval} reports results on a VRRPI-Bench test subset with a 70/30 class distribution. SKR-ID consistently improves performance over the corresponding base model. These results further demonstrate SKR-ID’s effectiveness under the tested class-distribution shift, rather than only when the test and training data have similar class distributions.

\begin{center}
\captionof{table}{Evaluation on the 70/30 imbalanced VRRPI-Bench test subset. All metrics are reported as percentages.}
\label{tab:imbalanced_eval}
\small
\begin{tabular}{llrrrrrrrr}
\toprule
Backbone & & F1 & Acc & Bal Acc & C1 F1 & C2 F1 & C1 Ratio & C2 Ratio & AUROC \\
\midrule
\multirow{2}{*}{LLaVA-OV} & Base & 41.9 & \textbf{69.8} & 50.1 & \textbf{82.1} & 1.6 & 99.3 & 0.7 & 50.1 \\
& SKR-ID & \textbf{53.2} & 55.5 & \textbf{55.6} & 63.5 & \textbf{43.0} & 52.0 & 48.0 & \textbf{57.0} \\
\multirow{2}{*}{Qwen3-VL} & Base & 30.0 & 32.0 & 47.1 & 16.3 & 43.4 & 9.8 & 87.6 & 48.3 \\
& SKR-ID & \textbf{46.2} & \textbf{46.5} & \textbf{51.8} & \textbf{50.2} & 42.2 & 37.5 & 62.5 & \textbf{54.4} \\
\multirow{2}{*}{VST-RL} & Base & 40.0 & 40.0 & 46.7 & 41.2 & 38.8 & 32.0 & 68.0 & 46.7 \\
& SKR-ID & \textbf{44.3} & \textbf{44.3} & \textbf{52.6} & \textbf{44.4} & \textbf{44.1} & 30.3 & 69.8 & \textbf{56.4} \\
\bottomrule
\end{tabular}%
\end{center}

\newpage
\input{checklist.tex}

\end{document}

%% file: tab/bias_analysis.tex
\begin{tabular}{lccccccccccccc}
\toprule
 & Base & Swap Opt. & Swap Img. \\
\midrule
 $\hat{y} \in \{\langle\textit{left}\rangle, \langle\textit{right}\rangle\}$ \\
 \midrule
 LLaVA-OV & 0.0\% / 100.0\% & 100.0\% / 0.0\% & 0.0\% / 100.0\% \\
 Qwen3-VL & 10.6\% / 89.4\% & 30.8\% / 69.2\% & 13.9\% / 86.1\%  \\
 VST-RL & 13.8\% / 86.1\% & 13.8\% / 86.1\% & 18.4\% / 81.5\% \\

 \midrule

 $\hat{y} \in \{\langle\textit{left}\rangle, \langle\textit{right}\rangle, \langle\textit{neither}\rangle\}$ \\
 \midrule

 LLaVA-OV & 0.0\% / 100.0\% & 95.6\% / 0.6\% & 0.0\% / 100.0\% \\
 Qwen3-VL & 11.9\% / 85.4\% & 26.5\% / 70.4\% & 15.9\% / 79.7\% \\
 VST-RL & 6.6\% / 92.7\% & 24.4\% / 75.2\% & 8.8\% / 91.1\% \\

\bottomrule
\end{tabular}

%% file: checklist.tex
\section*{NeurIPS Paper Checklist}

\begin{enumerate}

\item {\bf Claims}
    \item[] Question: Do the main claims made in the abstract and introduction accurately reflect the paper's contributions and scope?
    \item[] Answer: \answerYes{} 
    \item[] Justification: Our abstract and introduction clearly states the main contributions: 1. a comprehensive analysis on 3D spatial bias; 2. an annotation-free method SKR-ID, which helps mitigate the bias and re-express VLM's knowledge. The experimental results in Section~\ref{sec:vlm_bias} and Section~\ref{sec:results_discussions} support these claims.

\item {\bf Limitations}
    \item[] Question: Does the paper discuss the limitations of the work performed by the authors?
    \item[] Answer: \answerYes{} 
    \item[] Justification: Please refer to Section~\ref{sec:conclusion}.

\item {\bf Theory assumptions and proofs}
    \item[] Question: For each theoretical result, does the paper provide the full set of assumptions and a complete (and correct) proof?
    \item[] Answer: \answerNA{} 
    \item[] Justification: The paper does not contain formal theoretical results, theorems, or proofs. Our claims are supported through empirical evaluation and analysis.
    \begin{itemize}
        \item The answer \answerNA{} means that the paper does not include theoretical results. 
        \item All the theorems, formulas, and proofs in the paper should be numbered and cross-referenced.
        \item All assumptions should be clearly stated or referenced in the statement of any theorems.
        \item The proofs can either appear in the main paper or the supplemental material, but if they appear in the supplemental material, the authors are encouraged to provide a short proof sketch to provide intuition. 
        \item Inversely, any informal proof provided in the core of the paper should be complemented by formal proofs provided in appendix or supplemental material.
        \item Theorems and Lemmas that the proof relies upon should be properly referenced. 
    \end{itemize}

    \item {\bf Experimental result reproducibility}
    \item[] Question: Does the paper fully disclose all the information needed to reproduce the main experimental results of the paper to the extent that it affects the main claims and/or conclusions of the paper (regardless of whether the code and data are provided or not)?
    \item[] Answer: \answerYes{} 
    \item[] Justification: Please refer to Section~\ref{sec:experiment_settings} for details about the experimental settings.

\item {\bf Open access to data and code}
    \item[] Question: Does the paper provide open access to the data and code, with sufficient instructions to faithfully reproduce the main experimental results, as described in supplemental material?
    \item[] Answer: \answerYes{} 
    \item[] Justification: Please refer to Section~\ref{sec:task_forumulation} for the dataset and to our GitHub repository for the released code and reproduction instructions: \url{https://github.com/astudentuser/Codes_of_SKR-ID}.

\item {\bf Experimental setting/details}
    \item[] Question: Does the paper specify all the training and test details (e.g., data splits, hyperparameters, how they were chosen, type of optimizer) necessary to understand the results?
    \item[] Answer: \answerYes{} 
    \item[] Justification: Please refer to Section~\ref{sec:experiment_settings} for details about the experimental settings.

\item {\bf Experiment statistical significance}
    \item[] Question: Does the paper report error bars suitably and correctly defined or other appropriate information about the statistical significance of the experiments?
    \item[] Answer: \answerYes{} 
    \item[] Justification: Please refer to Section~\ref{sec:experiment_settings} for details about the experimental settings and Section~\ref{sec:results_discussions} for the results. We claim that all results are statistically significant and averaged over three runs.

\item {\bf Experiments compute resources}
    \item[] Question: For each experiment, does the paper provide sufficient information on the computer resources (type of compute workers, memory, time of execution) needed to reproduce the experiments?
    \item[] Answer: \answerYes{} 
    \item[] Justification: Please refer to Section~\ref{sec:experiment_settings} for details about computing resources used in the experiments.
    
\item {\bf Code of ethics}
    \item[] Question: Does the research conducted in the paper conform, in every respect, with the NeurIPS Code of Ethics \url{https://neurips.cc/public/EthicsGuidelines}?
    \item[] Answer: \answerYes{} 
    \item[] Justification: We have read the NeurIPS Code of Ethics and confirm that our research conforms to it in every respect.

\item {\bf Broader impacts}
    \item[] Question: Does the paper discuss both potential positive societal impacts and negative societal impacts of the work performed?
    \item[] Answer: \answerYes{} 
    \item[] Justification: Please refer to Section~\ref{sec:conclusion}.
    
\item {\bf Safeguards}
    \item[] Question: Does the paper describe safeguards that have been put in place for responsible release of data or models that have a high risk for misuse (e.g., pre-trained language models, image generators, or scraped datasets)?
    \item[] Answer: \answerNA{} 
    \item[] Justification: This paper does not have such risks.

\item {\bf Licenses for existing assets}
    \item[] Question: Are the creators or original owners of assets (e.g., code, data, models), used in the paper, properly credited and are the license and terms of use explicitly mentioned and properly respected?
    \item[] Answer: \answerYes{} 
    \item[] Justification: All the assets used in this paper are properly credited and the license and terms of use are explicitly mentioned and properly respected.

\item {\bf New assets}
    \item[] Question: Are new assets introduced in the paper well documented and is the documentation provided alongside the assets?
    \item[] Answer: \answerYes{} 
    \item[] Justification: The code for SKR-ID has been released on GitHub: \url{https://github.com/astudentuser/Codes_of_SKR-ID}.

\item {\bf Crowdsourcing and research with human subjects}
    \item[] Question: For crowdsourcing experiments and research with human subjects, does the paper include the full text of instructions given to participants and screenshots, if applicable, as well as details about compensation (if any)? 
    \item[] Answer: \answerNA{} 
    \item[] Justification: This paper does not involve crowdsourcing nor research with human subjects.

\item {\bf Institutional review board (IRB) approvals or equivalent for research with human subjects}
    \item[] Question: Does the paper describe potential risks incurred by study participants, whether such risks were disclosed to the subjects, and whether Institutional Review Board (IRB) approvals (or an equivalent approval/review based on the requirements of your country or institution) were obtained?
    \item[] Answer: \answerNA{} 
    \item[] Justification: This paper does not involve crowdsourcing nor research with human subjects.

\item {\bf Declaration of LLM usage}
    \item[] Question: Does the paper describe the usage of LLMs if it is an important, original, or non-standard component of the core methods in this research? Note that if the LLM is used only for writing, editing, or formatting purposes and does \emph{not} impact the core methodology, scientific rigor, or originality of the research, declaration is not required.
    \item[] Answer: \answerYes{} 
    \item[] Justification: This paper focuses on enhancing the spatial understanding of vision Large Language Models (LLMs).

\end{enumerate}